\documentclass[sigconf]{acmart}

\usepackage{colortbl}
\usepackage{cleveref}
\usepackage{diagbox} 
\usepackage{mathtools}

\DeclareMathOperator*{\argmax}{arg\,max}

\AtBeginDocument{%
  }

\copyrightyear{2025}
\acmYear{2025}
\setcopyright{cc}
\setcctype{by}
\acmConference[FAccT '25]{The 2025 ACM Conference on Fairness, Accountability, and Transparency}{June 23--26, 2025}{Athens, Greece}
\acmBooktitle{The 2025 ACM Conference on Fairness, Accountability, and Transparency (FAccT '25), June 23--26, 2025, Athens, Greece}
\acmDOI{10.1145/3715275.3732070}
\acmISBN{979-8-4007-1482-5/2025/06}

\begin{document}
\sloppy

\title[Perils of Label Indeterminacy]{Perils of Label Indeterminacy: A Case Study on Prediction of Neurological Recovery After Cardiac Arrest}

\author{Jakob Schoeffer}
\email{j.j.schoeffer@rug.nl}
\affiliation
{
  \institution{University of Groningen}
  \city{Groningen}
  \country{The Netherlands}
}
\orcid{0000-0003-3705-7126}

\author{Maria De-Arteaga}
\email{dearteaga@utexas.edu}
\affiliation
{
  \institution{University of Texas at Austin}
  \city{Austin}
  \state{TX}
  \country{USA}
}
\orcid{0000-0003-2297-3308}

\author{Jonathan Elmer}
\email{elmerjp@upmc.edu}
\affiliation
{
  \institution{University of Pittsburgh School of Medicine}
  \city{Pittsburgh}
  \state{PA}
  \country{USA}
}
\orcid{0000-0002-6648-5456}


\begin{abstract}
The design of AI systems to assist human decision-making typically requires the availability of labels to train and evaluate supervised models. Frequently, however, these labels are unknown, and different ways of estimating them involve unverifiable assumptions or arbitrary choices. In this work, we introduce the concept of \emph{label indeterminacy} and derive important implications in high-stakes AI-assisted decision-making. We present an empirical study in a healthcare context, focusing specifically on predicting the recovery of comatose patients after resuscitation from cardiac arrest. Our study shows that label indeterminacy can result in models that perform similarly when evaluated on patients with known labels, but vary drastically in their predictions for patients where labels are unknown. After demonstrating crucial ethical implications of label indeterminacy in this high-stakes context, we discuss takeaways for evaluation, reporting, and design.
\end{abstract}

\ccsdesc[500]{Computing methodologies~Machine learning}
\ccsdesc[500]{Information systems~Decision support systems}
\ccsdesc[500]{Human-centered computing~Human computer interaction (HCI)}

\keywords{Label indeterminacy, multiplicity, AI-assisted decision-making, healthcare, case study}

\maketitle

\section{Introduction}

AI systems are commonly used for decision support in high-stakes domains, such as healthcare.
This typically requires training of supervised AI models.
The evaluation of supervised AI models generally assumes access to ground-truth labels against which predictions or decisions can be assessed.
If we do not have access to such labels, this can undermine both training and evaluation of AI systems.
Unfortunately, ground-truth labels are not always available; they can be missing, incorrect, biased, or uncertain.
Research has studied the problem of missing labels~\citep{zschech2019prognostic,zhu2022introduction} as well as the risks of bias and errors contained in labels~\citep{jiang2020identifying,fogliato2020fairness}.
In this work, we tackle a related, pervasive, yet understudied problem: there is often an absence of a single, agreed-upon set of ground-truth labels.
Instead, labels are often estimated or selected through processes that involve unverifiable assumptions and arbitrary choices---which can have crucial implications.
We term this phenomenon \textit{label indeterminacy} and define it as follows: \textit{label indeterminacy refers to situations where practitioners cannot make a principled choice among different options to obtain labels, and as a result are faced with unresolvable multiplicity.}

There are multiple conditions that lead to label indeterminacy.
When building AI systems to support experts or automate decisions, it is frequently the case that historical observations are used to train them.
For instance, a bank may use historical data of loan repayment to train a model for assisting agents in determining who should receive a loan~\citep{bao2019integration}.
In this case, outcome labels are only observed conditioned on certain decisions.
Concretely, we only observe whether someone can repay a loan if that person was issued a loan in the first place.
For rejected applicants, we cannot determine what might have occurred if they had been granted the loan.
If we wanted to train an AI model that predicts whether someone will repay a loan, we cannot readily use data from rejected applicants because we do not have ground-truth labels for them~\citep{lakkaraju2017selective}.
To use such data for training, these instances could be ignored, counterfactual labels could be estimated, and both these choices could be paired with sampling bias correction methods.
However, choosing among these options, and executing any of them, involves making strong and often unverifiable assumptions~\citep{de2018learning}.
Alternatively, rather than relying on historical data, one may have access to labels that were purposefully collected for model training.
For instance, crowdsourcing of labels has been proposed for a variety of tasks, including toxicity detection in online content~\citep{gordon2022jury} or for medical relation extraction~\citep{dumitrache2018crowdsourcing}.
In such cases, it is frequently the case that multiple annotations are collected for each instance.
As a result, different ways of aggregating these annotations may result in different labels.

The machine learning literature has typically treated different sources of label indeterminacy as belonging to separate streams of research (e.g., missing labels vs. aggregating multiple labels). However, as we demonstrate in our work, these challenges can be intertwined in practice, since the arbitrary choices practitioners must make can lead to different problem formulations.
For instance, in settings where outcomes are selectively observed, one may choose to constrain oneself to observed labels and treat the problem as sampling bias, or collect expert assessments and treat it as a label aggregation problem.

In the present work, we study the implications of label indeterminacy for AI-assisted decision-making in high-stakes settings.
We do so via a case study in the healthcare domain, focusing on predicting patient recovery after cardiac arrest---a task that has received significant interest in critical care and resuscitation science~\citep{park2019prediction,hessulf2023predicting,kwon2019deep,mayampurath2022comparison}.
In this context, one needs to deal with indeterminate labels for patients who died after withdrawal of life-sustaining therapies.
This occurs when, based on clinical circumstances and individual judgments, a healthcare provider believes there is a poor prognosis for recovery, making continued aggressive care unlikely to confer benefit. 
We model ten different methods for handling indeterminate labels, based on approaches that have been used in the literature or that would be plausible to use.
We outline both the potential arguments in favor of their use and the unverifiable assumptions they necessitate.
The considered labels include both estimations based on historical decisions and estimations based on an international panel of experts in neurological prognostication.
We show that these ten different methods yield models that perform similarly on the subset of cases with known labels (i.e., where recovery is actually observed or patients died despite maximal support), while issuing fundamentally different predictions for patients with indeterminate labels.

Our findings have significant implications.
The machine learning literature has focused on proposing variants for counterfactual estimation and label aggregation, showing modeling improvements under certain assumptions.
However, as we show in our case study, in the real world these assumptions are often unverifiable or known to be violated.
As a result, system designers are forced to make arbitrary choices that have profound implications, raising ethical concerns: a different arbitrary choice can lead to differences in AI recommendations for decisions that have irreversible outcomes. 
Empirically, we find that in 19.6\% of patients in our testing sample with unknown labels, at least one method of handling indeterminate labels predicts no chance of recovery, while another method predicts at least a moderate recovery probability of over 10\%.
When these predictions guide clinical decisions regarding the continuation or withdrawal of life-sustaining therapies, the consequences are profound.

Moreover, we show that the perils of label indeterminacy are exacerbated by the fact that the different approaches may yield comparable performance when evaluated on patients with known labels, i.e., the subset of patients where performance can be directly assessed.
This means that these differences in predictions may go undetected, and that it may not be possible to discard approaches based on poor performance on the subset of instances with known labels.
As we argue in this work, this results in a form of multiplicity that undermines the viability of these models for use in high-stakes settings.
Thus, efforts to improve modeling of missing labels do not address a fundamental problem: the \emph{indeterminate} nature of these labels.
As a path forward, we advocate for novel sociotechnical designs that shift the objective of prediction to other constructs that do not suffer from indeterminacy but are still useful for humans to reach high-quality decisions.

\section{Background and Related Work}

\paragraph{Label Imputation and Sampling Bias Correction}

True labels can be missing for different reasons.
One reason of particular importance to our case study is the so-called \textit{selective labels problem}~\citep{lakkaraju2017selective,wei2021decision}.
Under selective labels, we only have a partial labeling of instances, and whether we observe true labels depends on historical decisions---such as whether or not life-sustaining therapies are withdrawn~\citep{de2023self}.
Other examples of missing labels include sample dropout in medical trials~\citep{hogan2004handling,little1996intent} or so-called \textit{positive and unlabeled data} (PU), where only certain labels are recorded~\citep{bekker2020learning,wang2025noise,perini2023learning}.

When ground-truth labels are unavailable for training, they must be directly or indirectly estimated~\citep{muller2021designing}. This is a problem that has received substantial attention in the statistics and machine learning literature.
Prior work has suggested different techniques for estimating counterfactuals, such as nearest-neighbor imputation~\citep{chen2000nearest}, which require assumptions regarding the relevance of a given distance metric.
Other works treat missing labels as a sampling bias problem and aim to correct for it.
This typically requires assuming that every case has a non-negligible probability of having a label observed~\citep{de2018learning} as well as that no unobservables affect the likelihood of observing a label~\citep{lakkaraju2017selective}, which are often unrealistic.
Another approach, which is ubiquitous in biomedical research~\citep{de2023self}, is to assume that historical decisions that led to not observing a true label were correct, and to impute the labels accordingly.
Training AI models on these labels can, however, result in predictions that mimic potentially erroneous historical decisions and lead to harmful self-fulfilling prophecies~\citep{de2023self,elmer2023time}.
Taken together, prior work has proposed many different approaches to impute missing labels and correct for sampling bias, albeit at the cost of making strong assumptions.
Through our case study, we show that choosing among the many available options often becomes an arbitrary choice, because each makes assumptions that are unverifiable or violated in practice.
As a result, claims of improved modeling abilities do not address the fundamental challenges encountered in high-stakes decision-making.

\paragraph{Label Aggregation} While in some cases labels are missing for some instances, in others there may be multiple labels for each instance.
Often, multiple estimates are collected for a single instance, with the goal of obtaining higher-quality labels~\citep{ipeirotis2014repeated,sheng2008get}.
A similar challenge emerges when convening expert panels, which are often considered the gold standard in generating high-quality labels~\citep{nguyen2015combining,willemink2020preparing}.
This raises the question of how to aggregate multiple assessments for a single instance, and how to deal with expert disagreement~\citep{uma2021learning,gordon2022jury,wallace2022debiased}.
For that purpose, different methods have been proposed, ranging from majority voting~\citep{quoc2013evaluation} to more involved techniques like jury learning~\citep{gordon2022jury}.
Crucially, if expert assessments correspond to fundamentally different worldviews, prior work has argued that reconciling them in a single aggregate value appears nonsensical~\citep{manski2016interpreting}.
In this work, we have the unique opportunity to make use of a dataset containing estimates from a panel of 38 world-leading experts in neurological prognostication to assess patient cases with respect to their recovery potential.
This allows us to study the real-world implications of different choices of aggregating this information in high-stakes decision-making when faced with label indeterminacy, compare the option of learning solely from historical data vs. integrating panel assessments, and highlight the ethical challenges these choices pose.

\paragraph{Construct Validity}
Prior work has noted the fact that available labels may suffer from \emph{construct validity issues}~\citep{jacobs2021measurement,passi2019problem}, i.e., a gap between the construct of interest and what is captured in proxies used as training labels~\citep{de2021leveraging}.
Some work has treated this as a statistical modeling challenge, aiming to ``bridge the construct gap''~\citep{de2021leveraging}.
Other work has noted that treating it as part of the problem formulation stage, and empirically assessing the implications of different alternatives, may guide the correct choice~\citep{obermeyer2019dissecting}.
Label indeterminacy may or may not arise as a result of construct validity issues.
For instance, the outcome observed may correspond to the construct of interest, but it may not always be available.
Furthermore, we place emphasis on the \emph{indeterminate} nature of this problem---meaning, it \emph{cannot} be determined.
We empirically show that making a principled choice among options may be impossible, forcing designers to grapple with the ethical implications of choosing arbitrarily.
Through this conceptual difference, we are able to elucidate a crucial problem in the design of AI systems for decision support, and propose a path forward through a sociotechnical lens.

\paragraph{Algorithmic Multiplicity}

Multiplicity is the phenomenon that different (potentially arbitrary) design choices in the development of AI systems can lead to different outcomes.
Various types of multiplicity have been proposed and empirically studied in prior work.
Predictive multiplicity refers to the situation where a prediction problem admits competing models with conflicting predictions~\citep{marx2020predictive,watson2023predictive,black2022model}.
In these cases, a single dataset can support multiple solutions that are equivalent on the basis of relevant established performance metrics.
\citet{brunet2022implications} highlight implications of multiplicity for explanations, and \citet{coston2021characterizing} for fairness.
All these notions of multiplicity have in common that they assume access to a single set of ground-truth labels.
This is an important conceptual difference to our work, where we have no grounds for determining what is correct for at least a portion of instances.

The degrees of freedom for system designers afforded by multiplicity bring about the possibility to prioritize other values in the model selection process besides predictive accuracy, but may also lead to problems around justifiability and cause loopholes where system designers can mask intentional discrimination~\citep{black2022model}.
In this work, we show that label indeterminacy can lead to a form of multiplicity when multiple possible avenues to construct labels are equally justifiable.
We empirically show that this may result in settings where AI predictions differ drastically for instances with unknown labels, while maintaining comparable performance for those with known labels.
This has critical implications for decision subjects and poses challenges for current practices in AI-assisted decision-making that have received limited attention in previous research.

\section{Label Indeterminacy}

AI systems for decision automation or support often rely on supervised machine learning algorithms, which require the availability of a ``ground-truth'' label $Y$.
In many applications, however, such a label is often the result of a process spanning problem formulation~\citep{passi2019problem} and data engineering~\citep{orr2024social}.
Obscured in the end result is the fact that many other possible labels could have emerged from that process, and arbitrary choices in formulation and estimation led to an implicit selection among them.
We introduce the term \textbf{\emph{label indeterminacy}} to refer to situations where at least a portion of the ground-truth labels are unknowable and different ways of estimating them involve unverifiable assumptions or arbitrary choices, rendering it impossible to make a principled selection among them.

Intertwined in this notion are two important components: one that relates to the properties of the world itself and one that relates to our statistical estimation capabilities.
The unknowable nature of the labels is a notion grounded in philosophy literature, which posits that indeterminacy is a property of the world itself rather than what we know about the world~\citep{richardson2024social}.
Hence, it is not something we can aim to resolve by acquiring new knowledge or information.
This refers to the label for a specific, individual instance.
For example, if life-sustaining therapies are withdrawn from a critically ill patient and as a result the patient dies, it can never be known whether that patient would have recovered had life-sustaining therapies been continued.
This unknowable attribute is present across domains, and it is an essential property of settings in which the question relates to counterfactuals: we can never know what might have occurred if events had unfolded differently.
In the context of machine learning, however, the quantity of interest is not the estimation for an instance in the past, but rather the prediction for an instance in the future.
In the example of estimating recovery, a label indicating what would have happened had we continued life-sustaining therapies is a means to an end: providing better prognosis for patients who have not yet been treated.
Consequently, we are interested in label estimation that yields reliable and accurate predictions for new patients.
The problem, however, emerges when multiple such labels exist, and it is impossible to make a principled choice among them, because they each make assumptions that are either themselves unknowable or known to be false.
In such cases, the quantity of interest is indeterminate.    

Formally, let $Y_i$ be a ground-truth label for instance $i \in I$, with $I=\{1,...,m\}$, $m \in \mathbb{N}$.
$Y_i$ can be known or unknown at the time of dataset construction.
If $Y_i$ is unknown, in some cases one can collect more empirical evidence to arrive at the true label; for instance, by means of human annotators~\citep{deng2014scalable}.
In other cases, $Y_i$ \emph{cannot} be known---which is the situation of interest in this work.
For these instances, there may be a set of $\ell$ plausible labels $\{Y_i^1, Y_i^2, ..., Y_i^{\ell}\}$, and knowing which among them is more likely to be correct may be impossible.
We refer to the subset of instances $W \subseteq I$ with unknowable labels as \textbf{\emph{label indeterminacy set}}.

Indeterminacy poses a fundamental problem for machine learning when the consequences of selecting one among the available labels deem it unethical to let this be an arbitrary choice.
Next, we empirically study label indeterminacy in a high-stakes setting, and derive implications for the development and deployment of AI systems for decision support. 

\section{Empirical Study}

\begin{figure*}[t]
    \centering
    \includegraphics[width=0.7\linewidth]{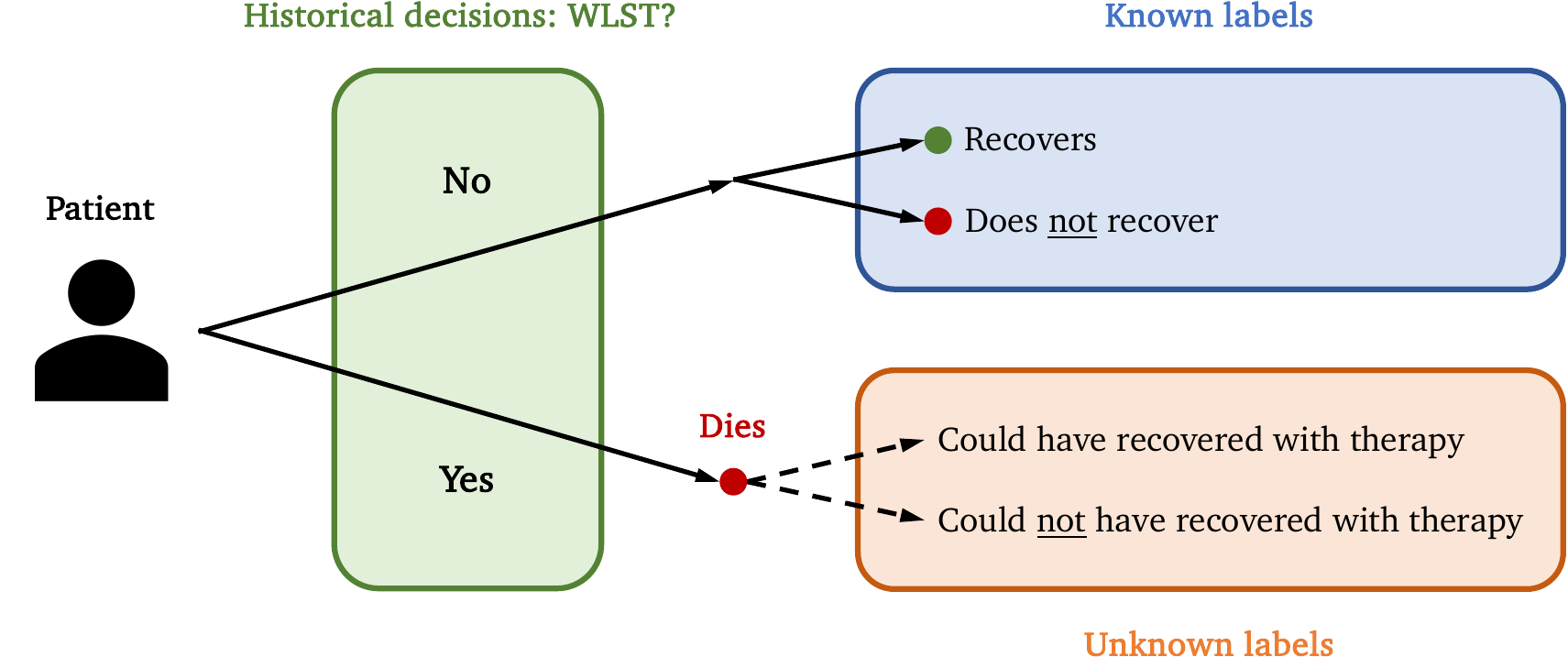}
    \caption{Obtaining labels depends on whether patients are withdrawn from life-sustaining therapies (WLST) or not. In the case of WLST, patients die, and we do not know whether they could have recovered had they not been withdrawn from life-sustaining therapies. For WLST cases, labels are unknown.
    }
    \Description{An illustration of label indeterminacy in the context of the case study.}
    \label{fig:selective_labels}
\end{figure*}

To assess the effects of label indeterminacy in practice, we conducted an empirical study in the healthcare domain, focused on predicting patient recovery after cardiac arrest.
In this section, we begin by presenting the use case and a summary of the data used in our study (\Cref{sec:use_case_data}).
In \Cref{sec:collecting_expert}, we describe the complementary data of expert assessments collected via a panel by \citet{elmer2025recovery}.
Following this, we explore various modeling approaches to deal with unknown labels (\Cref{sec:constructing_labels}) and compare these models' performance against the portion of available ground-truth labels (\Cref{sec:performance_observed}).
Finally, in \Cref{sec:differences_individual}, we compare the models' predictions for cases with indeterminate labels.

\subsection{Use Case and Historical Data}\label{sec:use_case_data}

\paragraph{Context}
Each year, over 150,000 Americans are hospitalized after resuscitation from cardiac arrest~\citep{martin20242024}.
Many of these patients are initially in a coma and require life-sustaining therapies in the intensive care unit; around 70\% do not survive their hospital stay~\citep{nolan2021european,callaway2015part}.
For comatose patients, the first major predictive task for doctors is typically to differentiate between patients who can awaken from coma (a necessary first step in recovery) and those for whom severe brain injury precludes awakening.
In the US, the surrogate decision-makers for most patients who are judged by doctors to have a good chance of recovery opt to continue life-sustaining therapies to facilitate this recovery.
Conversely, when doctors believe there is no chance of future awakening from coma, most decision-makers choose to \textbf{W}ithdraw \textbf{L}ife-\textbf{S}ustaining \textbf{T}herapies (\textbf{WLST}) in favor of a comfortable death.
Awakening from coma after cardiac arrest can take several days, and identifying potential for awakening is challenging, especially early on ~\citep{sandroni2014prognostication,paul2016delayed,nolan2021european}.
It is known that doctors are not perfectly accurate when making these prognostic judgments~\citep{steinberg2020providers,steinberg2022physicians}.
Misguided therapeutic nihilism~\citep{hemphill2009clinical} and knowledge gaps contribute to WLST and avoidable deaths of patients who would have recovered had life-sustaining treatments been continued.
Because death follows WLST decisions, recovery cannot be observed after WLST---this is illustrated in \Cref{fig:selective_labels}.

\paragraph{AI-Based Recovery Prediction}

Given the complexity of neurological prognosis, there is substantial interest in leveraging AI-based decision support for predicting recovery potential~\citep{park2019prediction,hessulf2023predicting,kwon2019deep,mayampurath2022comparison}.
AI systems, including supervised machine learning, offer a prospect of improving prognostication after cardiac arrest by providing more accurate and timely outcome predictions~\citep{kwon2019deep}.
The goal of most prognostic AI models is to predict recovery for individual patients based on their clinical characteristics.
These models are trained using existing data and seek to guide future medical care.
However, the presence of the selective labels problem poses the risk of creating harmful self-fulfilling prophecies, where flawed historical treatment decisions are repeated~\citep{de2023self}.
Hence, efforts have been made towards mitigating bias stemming from selective labels through statistical approaches~\citep{elmer2023time}.

\paragraph{Patient Data}

Data for our study are derived from a cohort of patients resuscitated from cardiac arrest who were hospitalized at a single US-based academic medical center.
Clinical information includes patient demographics, details of the circumstances of cardiac arrest, daily clinical summaries (e.g., neurological exam findings, prognostic test results, laboratory tests), and recovery (yes/no), which entails consciousness recovery (i.e., an ability to follow verbal commands) and survival to hospital discharge.
For our study, we consider a sample of 2,676 patients, in 1,026 of whom (38.3\%) we observe the true outcome (either recovered or not) because therapies were not withdrawn, and in 1,650 of whom (61.7\%) WLST results in unknowable labels.
For the subset of WLST cases, we collected expert assessments to gauge their counterfactual recovery potential for the hypothetical case that they had been provided with ongoing therapies.

\subsection{Panel Data}\label{sec:collecting_expert}

\paragraph{Panel Composition}
An international panel of 38 leading experts in neurological prognostication was recruited via their professional reputation~\citep{elmer2025recovery}.
To be included in the panel, experts had to meet at least two of the following three criteria.
They must have $(i)$ cared for over 150 post-arrest patients, $(ii)$ published more than three peer-reviewed articles on post-arrest care and prognostication in the past three years, and $(iii)$ played a leadership role in a major study of post-arrest care, such as a clinical trial.

\paragraph{Panel Annotation}
Over two years, IRB-approved one-hour case review sessions were conducted~\citep{elmer2025recovery} with groups of three or more experts via Zoom.
During each session, detailed clinical information was reviewed from individual patients who died after WLST.
Every session included approximately four cases, depending on the complexity of the cases.
Experts were presented with a structured clinical summary presentation and had an opportunity to ask clarifying questions.
The experts then navigated to the data collection instrument and independently entered their counterfactual outcome estimate quantifying the likelihood of recovery if life-sustaining therapies had been continued.
Experts estimated the counterfactuals along seven clinically relevant strata, depicted in \Cref{tab:strata}.

\begin{table}[ht]
    \centering
    \caption{7-point scale of expert assessments to predict the likelihood of patient recovery.}
    \label{tab:strata}
    \resizebox{\columnwidth}{!}{
    \begin{tabular}{c l c}
    \toprule
         \textbf{Expert rating} & \textbf{Interpretation} & \textbf{Likel. of recovery}  \\
         \midrule
         \textbf{0} &  \textit{``No chance of awakening''} & 0\% \\
         \textbf{1} &  \textit{``Trivial chance of awakening''} & (0\%, 1\%]\\
         \textbf{2} &  \textit{``Very small chance of awakening''} & (1\%, 5\%]\\
         \textbf{3} &  \textit{``Small chance of awakening''} & (5\%, 10\%]\\
         \textbf{4} &  \textit{``Moderate chance of awakening''} & (10\%, 25\%]\\
         \textbf{5} &  \textit{``Good chance of awakening''} & (25\%, 50\%]\\
         \textbf{6} &  \textit{``More likely than not to awaken''} & $>$50\%\\
         \bottomrule
    \end{tabular}
    }
\end{table}

Overall, 4,345 individual expert assessments were collected for WLST patients.
For our study, we only consider cases for which we have three assessments from different experts, leaving us with 1,428 cases.
All expert ratings from 0--6 were present in our data, and in many cases one or more experts believed there was substantial recovery potential.

\subsection{Different Ways of Handling Unknown Labels }\label{sec:constructing_labels}\label{sec:label_construction}

\begin{figure*}[t]
    \centering
    \includegraphics[width=0.75\linewidth]{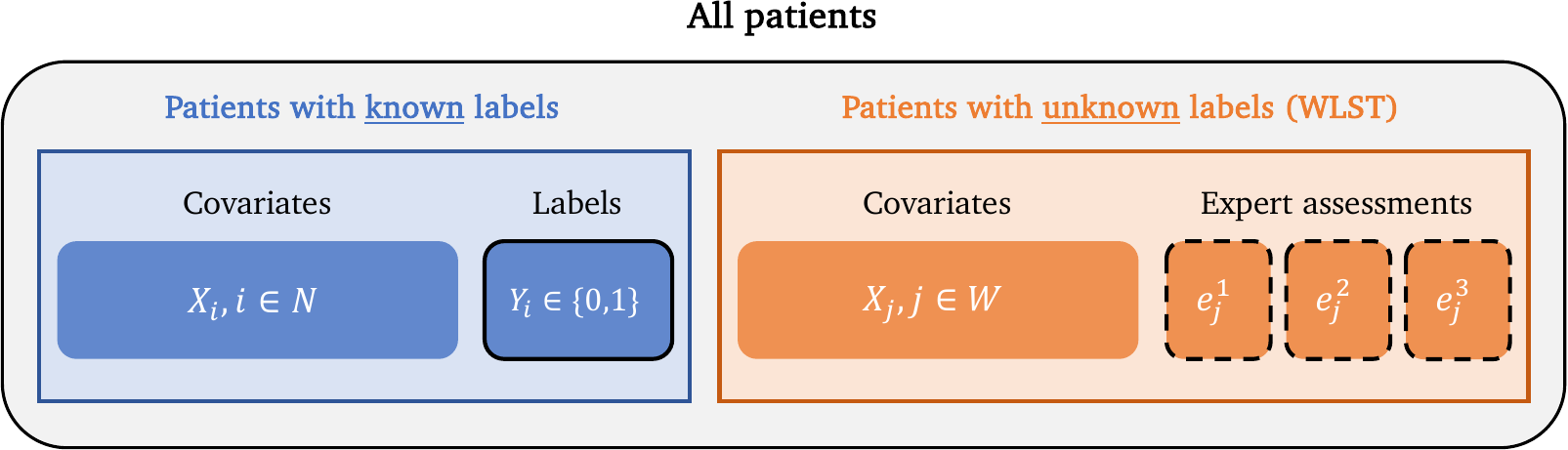}
    \caption{Data setup and taxonomy of our empirical study. The whole patient population is divided into patients \textit{with} (blue, left) and \textit{without} (orange, right) known labels. The subset of patients with known labels is $N$, and the label indeterminacy set (i.e., WLST patients) is $W$. For each WLST patient $j \in W$, we have three expert assessments, $e_j^1,e_j^2,e_j^3 \in [0,1]$. Dashed borders indicate that the corresponding values are \textit{not} ground truth.}
    \Description{Data setup and taxonomy of the case study. For some patients, there are ground-truth labels; for others (WLST), there are three expert assessments.}
    \label{fig:labels}
\end{figure*}

We outline the construction of ten predictive models, based on approaches to deal with indeterminate labels that have been used in the literature or that would be plausible to use.
For each, we include a brief discussion of motivations and unverifiable assumptions.
First, we establish some notation.
Let $X$ be the set of clinical information used as covariates.
Further, let $N$ be the set of patients with known labels $Y_i \in \{0,1\}$, $i \in N$, and let $W$ be the label indeterminacy set.
For any WLST patient $j\in W$, we have three expert assessments, $e_j^1,e_j^2,e_j^3 \in [0,1]$,\footnote{We translate discrete expert ratings to soft labels that correspond to the upper bound of the respective rating category. For instance, a rating of \textbf{3} corresponds to a predicted likelihood of recovery of 10\%.} which we treat as soft labels.\footnote{Note that for two WLST patients $j$ and $k$, with $j \neq k$, the expert ratings $e_j^\ell$ and $e_k^\ell$, $\ell \in\{1,2,3\}$, do not generally come from the same expert.}
The notation and data setup is summarized in \Cref{fig:labels}.

In the following, we detail different possible approaches to use this information for constructing models meant to predict a patient's likelihood of recovery.
In some cases, these approaches rely on estimating labels for WLST patients, which we denote $\widetilde{Y}_j \in [0,1]$; the tilde makes explicit that these labels are not ground truth.
We distinguish approaches that consider only historical data (\Cref{subsub:historical}) from ones that leverage expert assessments (\Cref{subsub:experts}).
We give models brief descriptive names and refer to them with $M$ and a corresponding subscript.

\subsubsection{Only Based on Historical Data}\label{subsub:historical}
Having access to a panel of experts who estimate counterfactuals for cases with unknown labels is a very rare occurrence in practice.
Hence, a number of methods can be employed to train a model using only available historical data.

\paragraph{Only observed ($M_{obs}$)}
Since we cannot know what would have happened for WLST cases, a common approach is to ignore WLST cases altogether and only use instances $i \in N$ with known labels for training.
However, this only yields a statistically unbiased model if labels are missing at random.
This assumption is not met in our domain because WLST patients differ fundamentally in their clinical characteristics and, as a result, WLST labels are \textit{not} missing at random.
Thus, relying only on cases with known labels risks learning from a different distribution than the target inference population, resulting in out-of-distribution predictions for patients that resemble WLST cases.

\paragraph{Only observed + IPW ($M_{obs+ip}$)}

A common approach to tackle the sampling bias in $M_{obs}$ is to apply inverse propensity weighting (IPW) during training~\citep{austin2015moving,little2019statistical}.
This is done by training a separate classifier that predicts the likelihood of observing the true label for each instance, and then weighing each instance $i \in N$ by their inverse probability of being observed.\footnote{In our empirical experiments, we use logistic regression for estimating the propensity scores, with a test AUC of .8199.}
This way, instances that resemble WLST cases receive high weight during training, correcting the sampling bias.
Note that in cases where the source of selection bias are historical expert decisions, this model is predicting those human decisions.
Crucially, this approach assumes that all instances have a non-negligible probability of
being part of the training sample---known as the \textit{positivity assumption}---and it also assumes that the decision that determines whether a label is observed does not depend on any unobserved confounders.
Both of these assumptions are hard to meet in practice~\citep{de2018learning}.
The assumption of no confounders is not one that can be verified solely from the data, while the positivity assumption is testable but frequently found to be violated.
In our data, neither of these assumptions hold; there are cases that have a negligible or null probability of non-withdrawal given that they present clear markers of poor neurological prognosis.
Moreover, WLST decisions are affected by factors not contained in the data, such as information communicated by the family members, as well as medical information that is not part of the available covariates $X$. 

\paragraph{Assume WLST correct ($M_{w_{corr}}$)}

Instead of ignoring WLST cases, this approach, which is ubiquitous in biomedical research~\citep{de2023self}, imputes $\widetilde{Y}_j = 0$ for all $j \in W$.
It then uses all cases that underwent WLST in addition to the ones with known labels for training.
The motivation for this approach is that WLST decisions are made by highly knowledgeable experts.
Hence, these decisions are often thought to be a reliable label.
While this approach ensures that WLST cases are represented, it risks erroneously associating factors that predict WLST with poor neurological outcome, because it assumes that historical decisions to withdraw life-sustaining therapies were correct.
In particular, if the motivation to use machine learning is to improve the quality of decisions, it is contradictory to assume that historical decisions were perfect.

\paragraph{Nearest-neighbor imputation ($M_{nn}$)}

An alternative way of estimating WLST labels is to impute them with labels from similar patients who were \textit{not} withdrawn from life-sustaining therapies.
A common method for achieving this is \textit{nearest-neighbor imputation}~\citep{chen2000nearest}.
In our study, we operationalize this for any $j \in W$ as follows: $\widetilde{Y}_j = Y_i$, with $i=\argmax_k \left(\frac{X_j \cdot X_k}{\Vert X_j \cdot X_k \Vert} \right)$ for $k \in N$.
This approach assumes knowledge of a relevant similarity metric, and for each WLST patient $j$, the presence of at least one patient $k \in N$ that is ``similar enough.''
The latter is equivalent to the positivity assumption, which we have discussed above.
The former is also typically violated, as standard metrics, such as Euclidean distance, are unlikely to capture relevant similarity, and attempts to learn a task-specific metric encounter the same data issues as attempts to predict neurological recovery.

\subsubsection{Leveraging Expert Assessments}\label{subsub:experts}
When multiple expert assessments are available for each case, the challenge is not how to deal with missing information, but rather how to aggregate multiple sources of information.
Naturally, when dealing with both missing historical data and multiple expert assessments, both issues co-exist.
We now discuss different ways in which one can leverage expert panel assessments to estimate labels for WLST cases, and ways of combining both types of approaches. 

\paragraph{Augment experts all ($M_{exp_{all}}$)}

A seemingly natural first way to construct labels from multiple expert assessments is to leverage every assessment individually.
This means that for each instance $j$ we have $k_j$ annotations, $e_j^1,...,e_j^{k_j}$.
In the training process of the respective model, each instance is represented $k_j$ times, each time with a potentially different label.
To account for this upsampling, the training instances are weighed by $\frac{1}{k_j}$.
In the context of our dataset, each WLST patient $j \in W$ has three labels, $\widetilde{Y}_j^1 = e_j^1$, $\widetilde{Y}_j^2 = e_j^2$, and $\widetilde{Y}_j^3 = e_j^3$, and is weighted by $\frac{1}{3}$.
This approach assumes that all expert assessments should be accounted for equally, even if it means training a model on potentially conflicting information.

\paragraph{Augment experts average ($M_{exp_{avg}}$)}

A prevalent practice in the literature on label aggregation~\citep{wei2023aggregate} is to take the mean of multiple assessments to arrive at one label per patient.
In our study, this means $\widetilde{Y}_j = \frac{1}{3} \cdot \left(e_j^1 + e_j^2 + e_j^3\right)$, $j \in W$.
In many cases, this approach yields a model that is equivalent to $M_{exp_{all}}$.
In this framing, however, useful information is removed from the data, and with it the potential to leverage that information by estimating quantities other than the most likely outcome.
In particular, expert disagreement might be based on fundamentally different worldviews, in which case the ``average expert'' is a meaningless measure~\citep{manski2016interpreting}.
Moreover, by aggregating, any potentially insightful variance in expert responses gets lost.

\paragraph{Augment experts max ($M_{exp_{max}}$)}
Instead of averaging expert assessments, another approach is to consider only the most optimistic prognoses.
The rationale is that if the goal is to counter therapeutic nihilism, then it should suffice to assign an optimistic label if just one expert identifies recovery potential.
In our study, this yields $\widetilde{Y}_j = \max \left(e_j^1,e_j^2,e_j^3 \right)$, $j \in W$.
This approach assumes that the expert determining the label is indeed right in their optimism, and that any potential disagreement with other experts is due to \textit{them} being overly pessimistic.

\paragraph{Augment expert agreement ($M_{exp_{agr}}$)}
Instead of augmenting \textit{all} WLST cases, the idea of this labeling approach is to only consider cases where experts agree in their assessments, and disregard those with substantial disagreement.
The intuition behind this is that it may be desirable to learn from experts when they exhibit consistency, as this may be indicative of underlying expert consensus, while excluding assessments with observed disagreement---especially if in such cases we can default to observed outcomes~\citep{de2021leveraging}.
We operationalize this in our study as follows:
\begin{equation*}
    \widetilde{Y}_j =
    \begin{dcases*}
        \max \left(e_j^1,e_j^2,e_j^3 \right) & \textbf{if}\ \ $\max \left(e_j^1,e_j^2,e_j^3 \right)\leq .01$ \textbf{or} \\ & $\min \left(e_j^1,e_j^2,e_j^3 \right) > .01$ \\
        \varnothing & \textbf{if}\ \ otherwise,
    \end{dcases*}
\end{equation*}
where we denote with $\varnothing$ labels that are not augmented.
This approach assumes that cases with expert disagreement carry no helpful information and, on the other hand, that expert agreement is indicative of correctness, which is an unverifiable assumption.

\begin{table*}[ht]
    \small
    \centering
    \caption{Means and standard deviations (STD) of AUC scores from 5-fold cross validation on patients with known labels. AUC scores are values between 0 and 1; higher AUC scores indicate better predictive performance.
    }
    \label{tab:auc}
    \begin{tabular}{l c c c c c c c c c c}
    \toprule
        & $M_{obs}$ & $M_{obs+ip}$ & $M_{w_{corr}}$ & $M_{nn}$ & $M_{exp_{all}}$ & $M_{exp_{avg}}$ & $M_{exp_{max}}$ & $M_{exp_{agr}}$ & $M_{exp_{agr_w}}$ & $M_{exp_{agr_w}+ip}$ \\
        \midrule
        Mean & .9247 & .9246 & .9249 & .9241 & .9269 & .9272 & .9260 & .9259 & .9244 & .9209 \\ 
        STD & .0220 & .0243 & .0175 & .0236 & .0152 & .0172 & .0159 & .0144 & .0222 & .0217 \\
        \bottomrule
    \end{tabular}
\end{table*}

\paragraph{Augment expert agreement on WLST ($M_{exp_{agr_w}}$)}
Augmenting only WLST cases where experts agree on the absence of recovery potential aims at mitigating the selective labels problem, which only affects our ability to observe the true outcome for WLST cases~\citep{de2018learning}.
Thus, the intuition is that we may only want to rely on expert assessments when we do not have the possibility of observing a true outcome.
We operationalize this as follows:
\begin{equation*}
    \widetilde{Y}_j =
    \begin{dcases*}
        \max \left(e_j^1,e_j^2,e_j^3 \right) & \textbf{if}\ \ $\max \left(e_j^1,e_j^2,e_j^3 \right)\leq .01$ \\
        \varnothing & \textbf{if}\ \ otherwise.
    \end{dcases*}
\end{equation*}
This approach also makes the unverifiable assumption that experts are correct whenever they are consistent.

\paragraph{Augment expert agreement on WLST + IPW ($M_{exp_{agr_w}+ip}$)}
This model combines the motivation of the previous model, $M_{exp_{agr_w}}$, with inverse propensity weighting (IPW) to address sampling bias~\citep{austin2015moving,little2019statistical}.
Here, it is important to note that cases where experts agree on WLST are precisely the ones that lead to a violation of the positivity assumption.
Recall that the positivity assumption is violated when we never observe an outcome for certain types of patients because, based on current medical knowledge, they are always withdrawn from life-sustaining therapies.
When the panel agreement reflects this, incorporating these labels allows us to address the violation of this assumption.
The approach first constructs the same WLST labels as $M_{exp_{agr_w}}$, treats them as ground truth, and then applies IPW during training.
It relies on the same implicit, unverifiable assumptions about expert consistency as before.

\subsection{Results: Predictions on Patients With Known Labels}\label{sec:performance_observed}

To evaluate the different approaches to deal with label indeterminacy introduced in \Cref{sec:constructing_labels}, we first evaluate the performance of the respective models on patients for whom we have known labels, i.e., patients that did not undergo WLST.
To do this, we set up a 5-fold cross-validation pipeline to make predictions on five disjoint holdout sets.
We use random forest regression models~\citep{breiman2001random}, which are commonly used in recovery prediction~\citep{mayampurath2022comparison}.
All ten models are trained on $(i)$ the cases with known labels (except for the holdout sets) plus $(ii)$ the respective WLST cases with estimated labels, if applicable (see \Cref{tab:summary_models} in Appendix~\ref{app:summary_models} for a summary).
As performance metric, we assess the area under the receiver operating characteristic (ROC) curve, short AUC, which is standard practice in medical diagnosis~\citep{huang2005using,elmer2023time}.
The AUC score is a measure between 0 and 1 that captures the predictive ability of a model.
We also qualitatively assess the ROC curves themselves.
ROC curves plot sensitivity (i.e., a model's ability to correctly identify patients with no recovery potential) against 1--specificity (i.e., the share of patients that are incorrectly predicted to have no recovery potential) for different decision thresholds~\citep{hoo2017roc}.

\begin{figure*}[ht]
    \centering
    \includegraphics[width=0.94\linewidth]{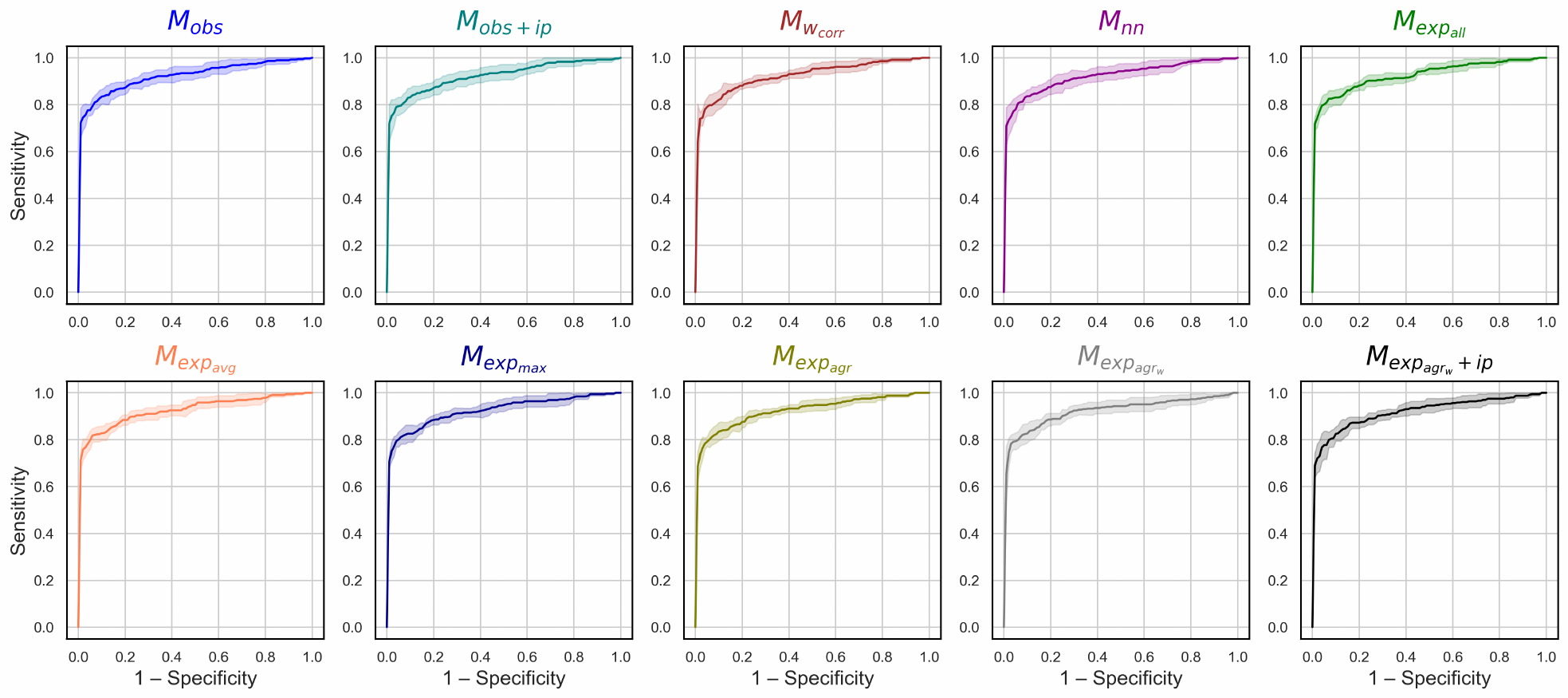}
    \caption{ROC curves with confidence bounds from 5-fold cross validation on patients with known labels. Shapes are nearly indistinguishable across models.}
    \Description{AUC curves for each model evaluated on patients with known labels. Shapes are nearly indistinguishable across models.}
    \label{fig:roc}
\end{figure*}

First, we note that predictive performance in terms of AUC is similar to prior studies on recovery prediction with AI, which lends clinical validity to our results~\citep{elmer2023time}.
Importantly, in \Cref{tab:auc}, we also see that the AUC scores are not significantly different between all ten models.
Moreover, \Cref{fig:roc} shows that sensitivity at different specificity levels is comparable across all models.
In other words, the predictive performances of all ten models are nearly indistinguishable on the subset of patients with known labels.
This shows that the choice of how to estimate and incorporate labels for WLST patients is arbitrary if we assess the different choices with respect to the resulting performance on patients with known labels.
We have ten different ways of handling indeterminate labels, each with unverifiable or knowingly violated assumptions, and they all exhibit statistically comparable performance when evaluated with respect to the portion of cases for which we know the true outcome.

\begin{figure*}[t]
    \centering
    \includegraphics[width=0.94\linewidth]{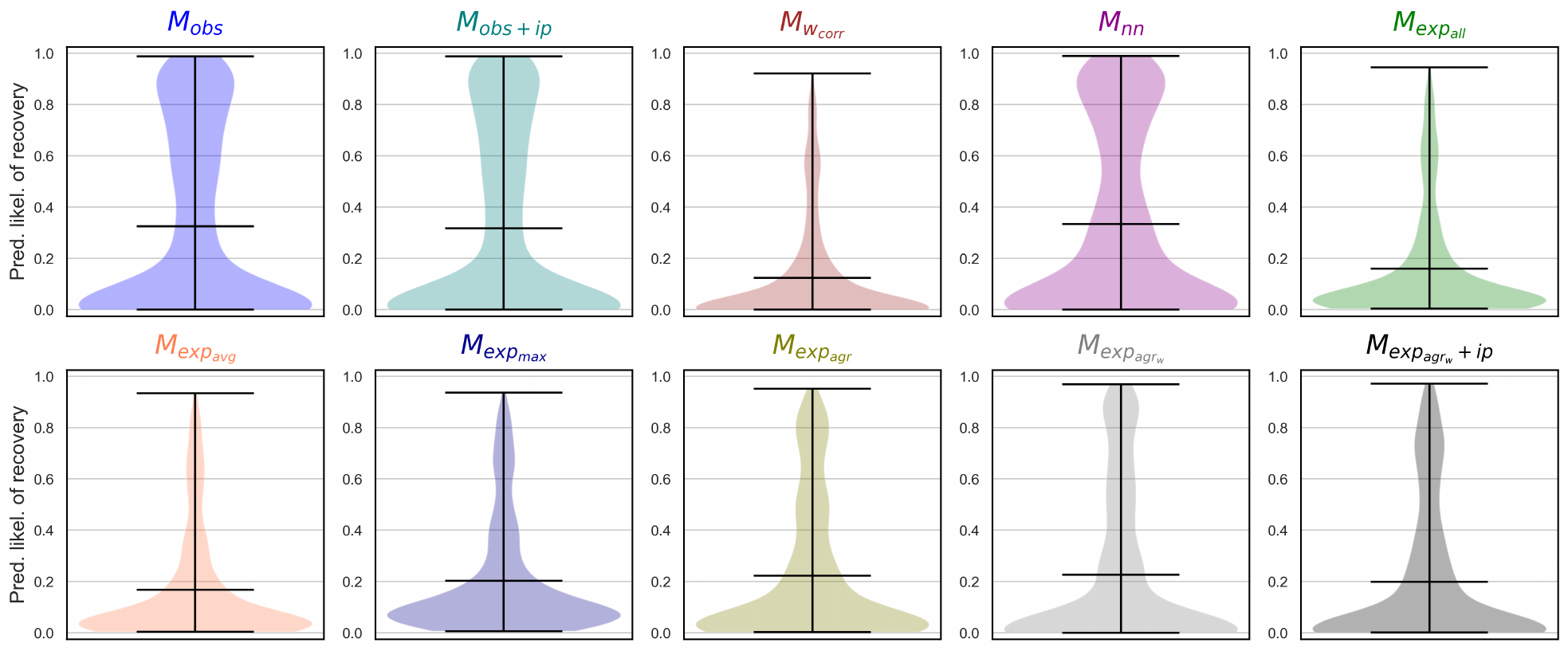}
    \caption{Distribution of predictions on holdout sets of WLST cases. Distributions are strikingly different across models, but we have no way of knowing which predictions are ``best'' in light of label indeterminacy.}
    \Description{Distribution of predictions for WLST patients by model. Distributions are strikingly different.}
    \label{fig:violins}
\end{figure*}

\subsection{Results: Predictions on Label Indeterminacy Set}\label{sec:differences_individual}

A similar evaluation is not possible for patients with indeterminate labels because of the absence of ground truth. 
But we are able to analyze differences in predictions across the different models.
Assuming a threshold of 1\% recovery likelihood---which is often used in clinical practice to inform decisions as to whether life-sustaining therapies should be continued or withdrawn~\citep{elmer2025recovery}---the following percentages of WLST patients would remain withdrawn under each model: 30.1\% ($M_{obs}$), 31.0\% ($M_{obs+ip}$), 52.0\% ($M_{w_{corr}}$), 23.4\% ($M_{nn}$), 6.7\% ($M_{exp_{all}}$), 6.9\% ($M_{exp_{avg}}$), 0.6\% ($M_{exp_{max}}$), 10.9\% ($M_{exp_{agr}}$), 43.9\% ($M_{exp_{agr_w}}$), 42.7\% ($M_{exp_{agr_w}+ip}$).
These stark differences indicate that label indeterminacy can have crucial practical implications.
We also see in \Cref{fig:violins} that the overall distributions of predictions for WLST patients vary significantly between models.

Intuitively, it seems plausible that the models which are only trained on known labels ($M_{obs}$ and $M_{obs+ip}$) may issue relatively more ``optimistic'' predictions, because the reason we observe true labels for them is that experts in the past were optimistic enough to not withdraw life-sustaining therapies.
Similarly, it appears sensible that the $M_{w_{corr}}$ model may issue relatively more pessimistic predictions if it learns that all WLST patients have no chance of recovery---and conversely, cases in which we augment the data by explicitly selecting the most optimistic expert assessments may yield models that are more optimistic.
Thus, based on the results shown in~\Cref{fig:violins}, it would be possible that models only vary in their calibration while maintaining the same ranked ordering of patients.
While this is typically not a big problem in machine learning, since probabilities can be calibrated via post-processing, in light of label indeterminacy, we would not be able to calibrate predictions because we have no grounds for knowing what the true outcome for WLST patients is.

\begin{table*}[ht]
    \small
    \centering
    \caption{Number of patients with predicted likelihood to recover of $\leq$1\% in one model (rows) and $>$10\% in another (columns). This shows that the ordering of predictions changes between models.}
    \label{tab:disagreement}
    \begin{tabular}{l c c c c c c c c c c}
    \toprule
    \diagbox{$\leq$1\%}{$>$10\%} & $M_{obs}$ & $M_{obs+ip}$ & $M_{w_{corr}}$ & $M_{nn}$ & $M_{exp_{all}}$ & $M_{exp_{avg}}$ & $M_{exp_{max}}$ & $M_{exp_{agr}}$ & $M_{exp_{agr_w}}$ & $M_{exp_{agr_w}+ip}$ \\
    \midrule
    $M_{obs}$ & \cellcolor{gray!25} & 0 & 0 & \textbf{76} & \textbf{4} & \textbf{2} & \textbf{35} & \textbf{7} & 0 & 0 \\
    $M_{obs+ip}$ & 0 & \cellcolor{gray!25} & 0 & \textbf{62} & \textbf{3} & \textbf{2} & \textbf{36} & \textbf{4} & 0 & \textbf{1} \\
    $M_{w_{corr}}$ & \textbf{111} & \textbf{83} & \cellcolor{gray!25} & \textbf{202} & \textbf{11} & \textbf{9} & \textbf{81} & \textbf{37} & \textbf{10} & \textbf{17} \\
    $M_{nn}$ & \textbf{25} & \textbf{10} & \textbf{1} & \cellcolor{gray!25} & \textbf{3} & \textbf{3} & \textbf{38} & \textbf{13} & \textbf{5} & \textbf{7} \\
    $M_{exp_{all}}$ & 0 & 0 & 0 & \textbf{7} & \cellcolor{gray!25} & 0 & 0 & 0 & 0 & 0 \\
    $M_{exp_{avg}}$ & 0 & 0 & 0 & \textbf{10} & 0 & \cellcolor{gray!25} & 0 & 0 & 0 & 0 \\
    $M_{exp_{max}}$ & 0 & 0 & 0 & \textbf{1} & 0 & 0 & \cellcolor{gray!25} & 0 & 0 & 0 \\
    $M_{exp_{agr}}$ & \textbf{1} & \textbf{1} & 0 & \textbf{16} & 0 & 0 & 0 & \cellcolor{gray!25} & 0 & 0 \\
    $M_{exp_{agr_w}}$ & \textbf{37} & \textbf{29} & 0 & \textbf{144} & \textbf{7} & \textbf{5} & \textbf{61} & \textbf{14} & \cellcolor{gray!25} & 0 \\
    $M_{exp_{agr_w}+ip}$ & \textbf{33} & \textbf{21} & 0 & \textbf{120} & \textbf{5} & \textbf{4} & \textbf{51} & \textbf{11} & 0 & \cellcolor{gray!25} \\
    \bottomrule
    \end{tabular}
\end{table*}

\begin{figure*}[h]
    \centering
    \includegraphics[width=0.87\linewidth]{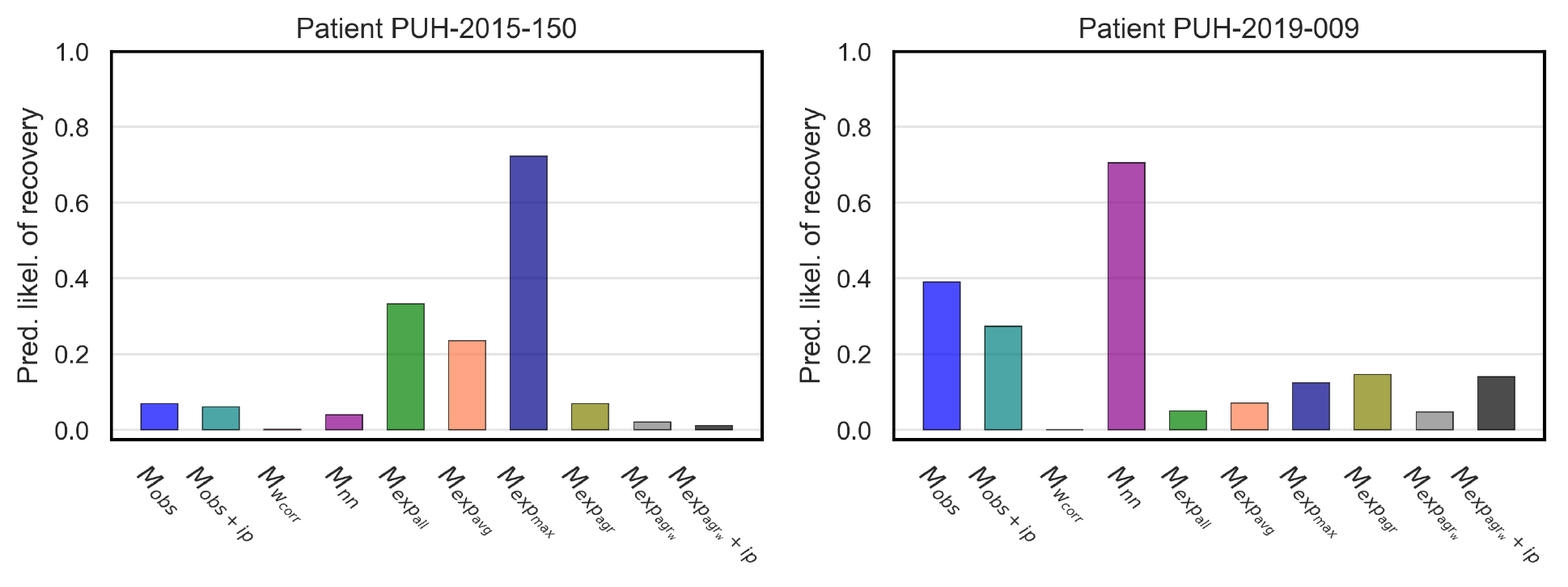}
    \caption{Predictions for two WLST patients with strong disagreement between models. This disagreement is entirely due to unverifiable design choices made in estimating and incorporating labels for WLST patients.}
    \Description{Predictions of all ten models for two different patients. Model disagreement is very high.}
    \label{fig:indiv}
\end{figure*}

Importantly, we find that the problem spans beyond calibration.
In our study, it is not the case that predictions are merely ``shifted'' up or down from one model to another; the ordering of patients also changes.
In other words, the difference between models is not only explained by different calibration.
For instance, there are 35 patients for whom the model $M_{exp_{max}}$ predicts a recovery potential of $>$10\% and $M_{obs}$ only $\leq$1\%. 
These pairwise comparisons are shown in \Cref{tab:disagreement} for all models, as well as in \Cref{tab:pairwise_25,tab:pairwise_50} in Appendix~\ref{app:additional_pairwise} for more extreme discrepancies.
Overall, 323 WLST patients (approx. 19.6\% of our sample) have a predicted likelihood to recover of $\leq$1\% by at least one model and $>$10\% by at least one other.
When comparing ranks,\footnote{We look at the ranking of patients to study whether the differences in predictions across models can be explained as a calibration issue, which we confirm is not the case. We acknowledge that the decision of whether to withdraw life-sustaining therapies is typically not a ranking problem---unless one is operating under limited resources, such as allocating respirators during the COVID-19 pandemic~\citep{white2020framework}.} 74 patients are ranked in the bottom third by at least one model and in the top third by at least one other model in terms of their likelihood to recover.
We visualize these stark differences in individual predictions in \Cref{fig:indiv} for two exemplary patients, where at least one model predicts a near-zero likelihood to recover and at least one other predicts $\gg$50\%.
In particular, the models providing optimistic vs. pessimistic predictions can vary significantly depending on the patient.
For instance, $M_{exp_{max}}$ estimates a recovery chance of over 70\% for patient PUH-2015-150 (\Cref{fig:indiv} left), while $M_{nn}$ suggests a recovery likelihood of less than 5\% for the same patient.
In contrast, when we examine patient PUH-2019-009 (\Cref{fig:indiv} right), the predictions are reversed.
These differences---which have critical and potentially irreversible implications for patients---are entirely a result of unverifiable design choices made in handling indeterminate labels.

Crucially, when comparing the heterogeneity of model predictions between WLST and non-WLST (i.e., the ones where labels are known) patients, we infer from \Cref{fig:boxplots-comparison,fig:barplots-comparison} in Appendix~\ref{app:comparison-multiplicity} that models tend to disagree more for WLST patients---both in terms of prediction disparities and the number of patients receiving conflicting recovery prognoses.
This shows that the disagreement we observe across models is not simply a function of predictive multiplicity.
Moreover, the increased heterogeneity for WLST patients is especially problematic because the WLST cases are precisely the ones for which we do \textit{not} have true labels and, as a result, cannot assess the correctness of predictions.
This results in fundamental limitations for principled model selection.

\section{Implications}

\paragraph{Ethical}

Deploying AI systems in high-stakes domains introduces specific ethical challenges that stem from label indeterminacy.
A key concern is the arbitrariness of decisions these systems may inform, wherein different and unjustifiable methodological choices for handling indeterminate labels can yield outcomes lacking any principled grounding~\citep{creel2022algorithmic,black2022model}.
For instance, a system might predict no chance of recovery for a patient under one label-handling method, while another method suggests a positive recovery potential for that same patient.
In a healthcare context, this arbitrariness may influence whether life-sustaining therapies are withdrawn, potentially resulting in premature life-or-death decisions.
This echoes concerns stemming from predictive multiplicity~\citep{marx2020predictive,d2022underspecification}.
While there is a similarity insofar as arbitrary choices yield different predictions, there is a differentiating factor: in the case of label indeterminacy, the issue does not stem from underspecification, under which one problem formulation with a known true label accepts multiple solutions~\citep{d2022underspecification}.
Instead, it stems from an ill-posed problem formulation, which relies on unknowable information, and in the face of it makes an arbitrary choice as to what that information may be.
The ``ground truth'' is chosen to be one among many options, without knowledge of which one is more likely to be correct, but with full knowledge that at most one of them may be correct.
As a result, the differences in predictions stem from some models relying on incorrect information.
This poses unique ethical challenges for the design and deployment of these systems.

Crucially, in current practice there is often an implicit and unchallenged assumption that one method of estimating labels is correct, with alternative approaches rarely compared.
This narrow focus obfuscates the indeterminate nature of the problem.
Moreover, the opacity of assumptions implicit in this selection process can obscure the mechanisms behind AI recommendations.
For patients and providers, this is particularly problematic as it veils an important dimension of uncertainty.
This can, among other risks, foster an over-reliance on these recommendations~\citep{buccinca2021trust,schoeffer2023interdependence}.
Without a clear understanding of the uncertainties and limitations, patients and providers might trust the AI recommendations uncritically, potentially introducing new and consequential risks in care delivery.
We refer to \citet{mertens2021chasing}, among others, for additional ethical concerns related to the use of AI systems in neurological prognostication, which are beyond the focus of this work.

\paragraph{Evaluation}
Label indeterminacy presents significant challenges for evaluating AI systems and AI-assisted decision-making.
Traditional machine learning evaluation methods often rely on the accuracy of predictions relative to the training labels, but when labels themselves are indeterminate, this approach becomes fallible.
Predictive accuracy alone is insufficient for assessing AI systems under these conditions, as it does not account for the inherent uncertainty of the labels.
Consequently, interpretations of studies that claim to measure accuracy or fairness in systems trained on (partially) indeterminate labels should be approached with caution, given the strong assumptions they necessarily make.

This also carries implications for research and practice on human-AI collaboration~\citep{fok2023search,bansal2021does,schemmer2023appropriate}.
Recent work in this field has advocated for a focus on \textit{appropriate reliance} when evaluating AI-assisted decision-making, focusing on enabling humans to override mistaken AI advice~\citep{schemmer2023appropriate,he2023knowing,schoeffer2024explanations,fok2023search}.
However, the proposed metrics are not readily applicable in the face of label indeterminacy, where the correctness of predictions is unknowable.
In light of this, we suggest that our evaluation approach can serve as a blueprint for future studies: $(i)$ split the data into the set of instances with known labels and the label indeterminacy set; $(ii)$ evaluate models through traditional performance metrics, such as accuracy or AUC, on the set of instances with known labels (if any); $(iii)$ compare differences in predictions on the label indeterminacy set.
An important open question for future work is how to define notions of appropriate reliance in the presence of label indeterminacy.

\paragraph{Reporting}

Reporting practices for AI-assisted decision-making typically contain an account of ``what the AI system is capable of doing''~\citep{amershi2019guidelines}. 
Making indeterminacy explicit in reporting would significantly improve transparency in the deployment of AI systems for decision support by offering a clearer picture of their limitations.
Additionally, when true labels are not readily available and are instead \emph{constructed}, it is crucial to report the choices and assumptions underlying this process.
In particular, we recommend reporting $(i)$ \emph{methods} used as part of the label construction, such as different forms of label aggregation or counterfactual estimation; $(ii)$ \emph{motivation} or rationale underlying the method choice; $(iii)$ \emph{assumptions} the method makes, and whether these are verifiable and indeed verified; and $(iv)$ \emph{arbitrary choices} made throughout the construction process, such as uninterpretable parameters, which may impact the resulting labels and, thus, predictions.
These dimensions extend existing reporting frameworks such as data cards \citep{pushkarna2022data}, model cards \citep{mitchell2019model}, or datasheets for datasets \citep{gebru2021datasheets}.
When multiple methods for label construction are considered, these should all be reported, together with their implications for the resulting tools and corresponding performances.

\paragraph{Design}

When designing AI systems for decision support, it is crucial to reconsider both the nature of AI predictions and their usage.
Traditionally, AI systems focus on predicting specific outcomes, but our work shows how unreliable this approach is in the light of label indeterminacy.
We encourage an increased focus on sociotechnical designs: rather than attempting to predict the outcome itself (e.g., recovery or loan default), one could design AI systems that predict a piece of information not suffering from indeterminacy and that is useful---in combination with other complementary sources of information---to reach a high-quality decision~\citep{holstein2023toward}.
For instance, AI systems could provide insights into how experts have evaluated similar cases in the past~\citep{ehsan2021expanding}.
On these grounds, our work also has important implications for human-centered explainable AI~\citep{ehsan2022human} and research on cognitive forcing functions~\citep{buccinca2021trust}.
Rather than focusing solely on elucidating the ``inner workings of AI,'' research in these areas should prioritize mechanisms that incorporate social dimensions of decision-making~\citep{smart2024beyond}.
Information concerning what construct is being predicted, and its differences with the construct of interest to the task, should be clearly communicated to users; otherwise, it could lead to a drift in decision-making criteria~\citep{green2019principles}.

Moreover, incorporating uncertainty inherent to indeterminacy into predictions could potentially enhance decision support; for instance, through informing deferrals or second opinions~\citep{lu2024does,madras2018predict,mozannar2023should}.
Instead of offering only a point prediction, systems might predict uncertainty based on whether experts have previously reached consensus on similar cases, or the degree to which individual predictions differ based on arbitrary design choices.
Uncertainty is considered a key concept in decision support~\citep{leibig2017leveraging,fernandes2018uncertainty,zhang2024rethinking}, and label indeterminacy provides a novel perspective for developing techniques to estimate and communicate this uncertainty.

\section{Conclusion}

We introduced the concept of \textit{label indeterminacy}, which refers to the fact that labels for training and evaluating AI systems are often unknowable, and handling them involves making unverifiable assumptions or arbitrary choices.
We conducted a comprehensive empirical study on the implications of label indeterminacy in the high-stakes scenario of predicting recovery of comatose patients after cardiac arrest.
Our findings highlight that different, unverifiable choices of estimating and incorporating labels can result in significantly different AI predictions, which may influence decisions with irreversible consequences.
Importantly, these differences might go unnoticed if model performance is evaluated solely on instances with known labels---as is commonly done.
This carries profound ethical implications and underscores the need for careful consideration in evaluation, reporting, and design.

\begin{acks}
This research was supported by NIH/NHLBI grant R01NS124642, by a Google Award for Inclusion Research, and by \textit{Good Systems},\footnote{\url{http://goodsystems.utexas.edu/}} a UT Austin Grand Challenge to develop responsible AI technologies.
The authors acknowledge the efforts of the Optimizing Recovery Prediction After Cardiac Arrest (ORCA) Study Group participants, who reviewed each case when death occurred after limitation or withdrawal of life-sustaining therapies to provide estimates of recovery potential had life-sustaining treatments been continued: Dr.~Alain Cariou, Dr.~Alejandro A. Rabinstein, Dr.~Alexis Steinberg, Dr.~Andrea O. Rossetti, Dr.~Ankur A. Doshi, Dr.~Bradley J. Molyneaux, Dr.~Cameron Dezfulian, Dr.~Carolina B. Maciel, Dr.~Cecelia Ratay, Dr.~Christoph Leithner, Dr.~Cindy Hsu, Dr.~Claudio Sandroni, Dr.~Clifton W.  Callaway, Dr.~David M. Greer, Dr.~David B. Seder, Dr.~Francis X. Guyette, Dr.~Fabio Silvio Taccone, Dr.~Hiromichi Naito, Dr.~Jasmeet Soar, Dr.~Jean-Baptiste Lascarrou, Dr.~Jerry P. Nolan, Dr.~Jonathan Elmer, Dr.~Karen G. Hirsch, Dr.~Katherine Berg, Dr.~Marion Moseby-Knappe, Dr.~Markus B. Skrifvars, Dr.~Michael Donnino, Dr.~Michael Kurz, Dr.~Min Jung Kathy Chae, Dr.~Mypinder Sekhon, Dr.~Nicholas J. Johnson, Mr.~Patrick J. Coppler, Dr.~Pedro Kurtz, Dr.~Romergryko G. Geocadin, Dr.~Sachin Agarwal, Dr.~Teresa L. May, and Dr.~Theresa Mariero Olasveengen.
\end{acks}

\bibliographystyle{ACM-Reference-Format}
\bibliography{references}

\onecolumn
\appendix

\section{Summary of All Models in Our Study}\label{app:summary_models}

\Cref{tab:summary_models} summarizes the ten different approaches to estimating and incorporating unknown labels for WLST cases in our study. 
Recall that $\widetilde{Y}_j$ denotes estimated labels, and $\varnothing$ refers to situations where labels are not augmented.

\begin{table*}[ht]
    \centering
    \caption{Summary of all models and their respective label construction for WLST cases.}
    \label{tab:summary_models}
    \begin{tabular}{l l}
    \toprule
    \textbf{Models} & \textbf{Label construction for WLST cases $j \in W$}  \\
    \midrule
    $M_{obs}$, $M_{obs+ip}$ & $\widetilde{Y}_j = \varnothing$  \\[5pt]
    $M_{w_{corr}}$ & $\widetilde{Y}_j = 0$ \\[5pt]
    $M_{nn}$ & $\widetilde{Y}_j = Y_i$, with $i=\argmax_k \left(\frac{X_j \cdot X_k}{\Vert X_j \cdot X_k \Vert} \right)$ for $k \in N$ \\
    \midrule
    $M_{exp_{all}}$ & $\widetilde{Y}_j^1 = e_j^1$, $\widetilde{Y}_j^2 = e_j^2$, $\widetilde{Y}_j^3 = e_j^3$ \\[5pt]
    $M_{exp_{avg}}$ & $\widetilde{Y}_j = \frac{1}{3} \cdot (e_j^1 + e_j^2 + e_j^3)$ \\[5pt]
    $M_{exp_{max}}$ & $\widetilde{Y}_j = \max (e_j^1,e_j^2,e_j^3 )$ \\[5pt]
    $M_{exp_{agr}}$ & $\widetilde{Y}_j =
    \begin{dcases*}
        \max (e_j^1,e_j^2,e_j^3 ) & \textbf{if}\ \ $\max (e_j^1,e_j^2,e_j^3 )\leq .01$ \textbf{or} $\min (e_j^1,e_j^2,e_j^3 ) > .01$ \\
        \varnothing & \textbf{if}\ \ otherwise
    \end{dcases*}$ \\[5pt]
    $M_{exp_{agr_w}}$, $M_{exp_{agr_w}+ip}$ & $\widetilde{Y}_j =
    \begin{dcases*}
        \max (e_j^1,e_j^2,e_j^3 ) & \textbf{if}\ \ $\max (e_j^1,e_j^2,e_j^3 )\leq .01$ \\
        \varnothing & \textbf{if}\ \ otherwise
    \end{dcases*}$\\
    \bottomrule
    \end{tabular}
\end{table*}

\section{Additional Pairwise Comparison Tables}\label{app:additional_pairwise}

\Cref{tab:pairwise_25,tab:pairwise_50} show the number of WLST patients for whom one model (rows) predicts a low likelihood of recovery ($\leq$1\%) and a different model (columns) predicts a relatively high likelihood of recovery ($>$25\% in \Cref{tab:pairwise_25}, and $>$50\% in \Cref{tab:pairwise_50}).

\begin{table*}[ht]
    \centering
    \caption{Number of WLST patients with predicted likelihood to recover of $\leq$1\% by one model (rows) and $>$25\% by another (columns).}
    \label{tab:pairwise_25}
    \begin{tabular}{l c c c c c c c c c c}
    \toprule
    \diagbox{$\leq$1\%}{$>$25\%} & $M_{obs}$ & $M_{obs+ip}$ & $M_{w_{corr}}$ & $M_{nn}$ & $M_{exp_{all}}$ & $M_{exp_{avg}}$ & $M_{exp_{max}}$ & $M_{exp_{agr}}$ & $M_{exp_{agr_w}}$ & $M_{exp_{agr_w}+ip}$ \\
    \midrule
    $M_{obs}$ & \cellcolor{gray!25} & 0 & 0 & \textbf{19} & 0 & 0 & \textbf{1} & \textbf{2} & 0 & 0 \\
    $M_{obs+ip}$ & 0 & \cellcolor{gray!25} & 0 & \textbf{15} & 0 & 0 & \textbf{2} & \textbf{2} & 0 & 0 \\
    $M_{w_{corr}}$ & \textbf{29} & \textbf{14} & \cellcolor{gray!25} & \textbf{101} & \textbf{2} & 0 & \textbf{6} & \textbf{5} & \textbf{6} & \textbf{6} \\
    $M_{nn}$ & \textbf{7} & 0 & 0 & \cellcolor{gray!25} & 0 & 0 & \textbf{3} & \textbf{3} & \textbf{2} & \textbf{1} \\
    $M_{exp_{all}}$ & 0 & 0 & 0 & 0 & \cellcolor{gray!25} & 0 & 0 & 0 & 0 & 0 \\
    $M_{exp_{avg}}$ & 0 & 0 & 0 & 0 & 0 & \cellcolor{gray!25} & 0 & 0 & 0 & 0 \\
    $M_{exp_{max}}$ & 0 & 0 & 0 & 0 & 0 & 0 & \cellcolor{gray!25} & 0 & 0 & 0 \\
    $M_{exp_{agr}}$ & 0 & 0 & 0 & \textbf{1} & 0 & 0 & 0 & \cellcolor{gray!25} & 0 & 0 \\
    $M_{exp_{agr_w}}$ & \textbf{5} & \textbf{4} & 0 & \textbf{56} & \textbf{1} & 0 & \textbf{4} & \textbf{5} & \cellcolor{gray!25} & 0 \\
    $M_{exp_{agr_w}+ip}$ & \textbf{3} & \textbf{1} & 0 & \textbf{44} & 0 & 0 & \textbf{3} & \textbf{4} & 0 & \cellcolor{gray!25} \\
    \bottomrule
    \end{tabular}
\end{table*}

\begin{table*}[ht]
    \centering
    \caption{Number of WLST patients with predicted likelihood to recover of $\leq$1\% by one model (rows) and $>$50\% by another (columns).}
    \label{tab:pairwise_50}
    \begin{tabular}{l c c c c c c c c c c}
    \toprule
    \diagbox{$\leq$1\%}{$>$50\%} & $M_{obs}$ & $M_{obs+ip}$ & $M_{w_{corr}}$ & $M_{nn}$ & $M_{exp_{all}}$ & $M_{exp_{avg}}$ & $M_{exp_{max}}$ & $M_{exp_{agr}}$ & $M_{exp_{agr_w}}$ & $M_{exp_{agr_w}+ip}$ \\
    \midrule
    $M_{obs}$ & \cellcolor{gray!25} & 0 & 0 & \textbf{3} & 0 & 0 & 0 & 0 & 0 & 0 \\
    $M_{obs+ip}$ & 0 & \cellcolor{gray!25} & 0 & \textbf{2} & 0 & 0 & 0 & 0 & 0 & 0 \\
    $M_{w_{corr}}$ & \textbf{2} & 0 & \cellcolor{gray!25} & \textbf{40} & 0 & 0 & \textbf{2} & 0 & 0 & 0 \\
    $M_{nn}$ & 0 &0  & 0 & \cellcolor{gray!25} & 0 & 0 & 0 & 0 & 0 & 0 \\
    $M_{exp_{all}}$ & 0 & 0 & 0 & 0 & \cellcolor{gray!25} & 0 & 0 & 0 & 0 & 0 \\
    $M_{exp_{avg}}$ & 0 & 0 & 0 & 0 & 0 & \cellcolor{gray!25} & 0 & 0 & 0 & 0 \\
    $M_{exp_{max}}$ & 0 & 0 & 0 & 0 & 0 & 0 & \cellcolor{gray!25} & 0 & 0 & 0 \\
    $M_{exp_{agr}}$ & 0 & 0 & 0 & 0 & 0 & 0 & 0 & \cellcolor{gray!25} & 0 & 0 \\
    $M_{exp_{agr_w}}$ & 0 & 0 & 0 & \textbf{18} & 0 & 0 & \textbf{1} & 0 & \cellcolor{gray!25} & 0 \\
    $M_{exp_{agr_w}+ip}$ & 0 & 0 & 0 & \textbf{10} & 0 & 0 & 0 & 0 & 0 & \cellcolor{gray!25} \\
    \bottomrule
    \end{tabular}
\end{table*}

\section{Prediction Heterogeneity for WLST vs. Non-WLST Patients}\label{app:comparison-multiplicity}

\begin{figure*}
    \centering
    \includegraphics[width=0.91\linewidth]{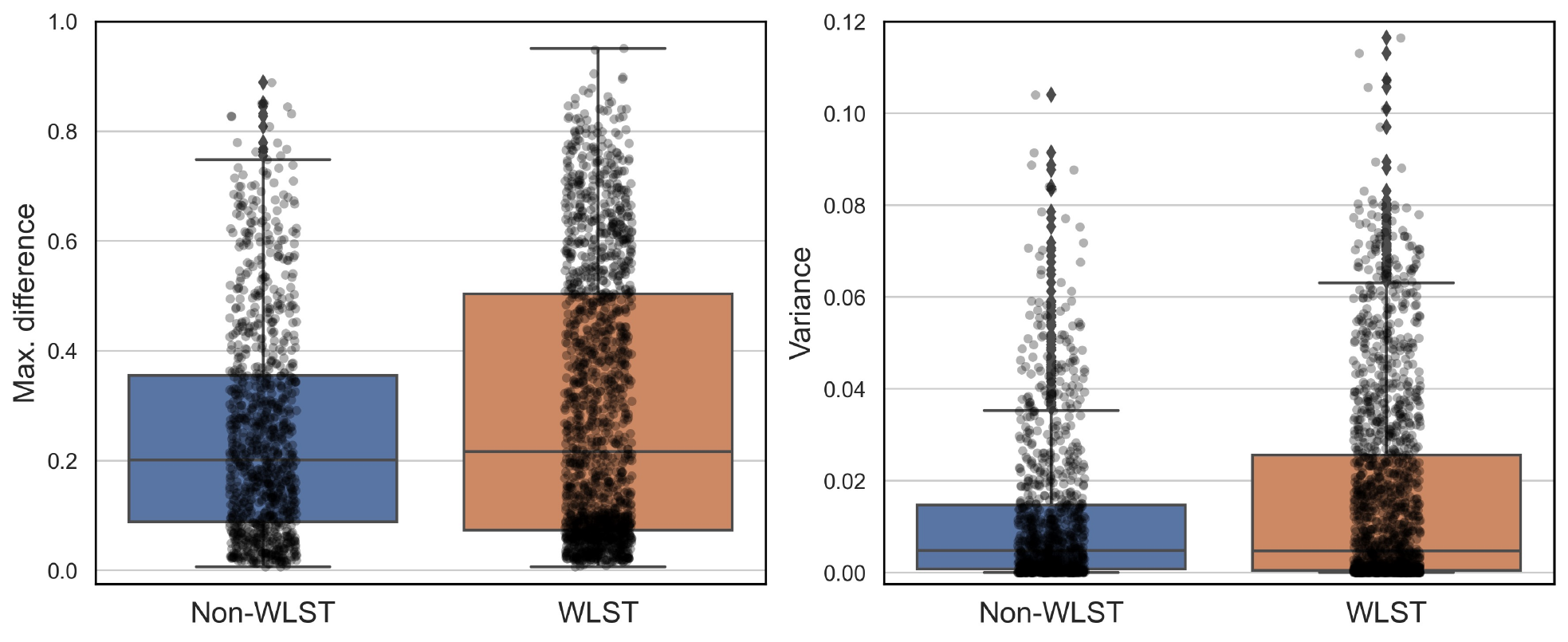}
    \caption{Strip and box plots of maximum difference (left) and variance (right) between different models' predictions for non-WLST (blue) and WLST (orange) patients.}
    \Description{Plots of maximum difference and variance between model predictions for non-WLST and WLST patients. Both metrics are higher for WLST patients.}
    \label{fig:boxplots-comparison}
\end{figure*}

\begin{figure*}
    \centering
    \includegraphics[width=0.84\linewidth]{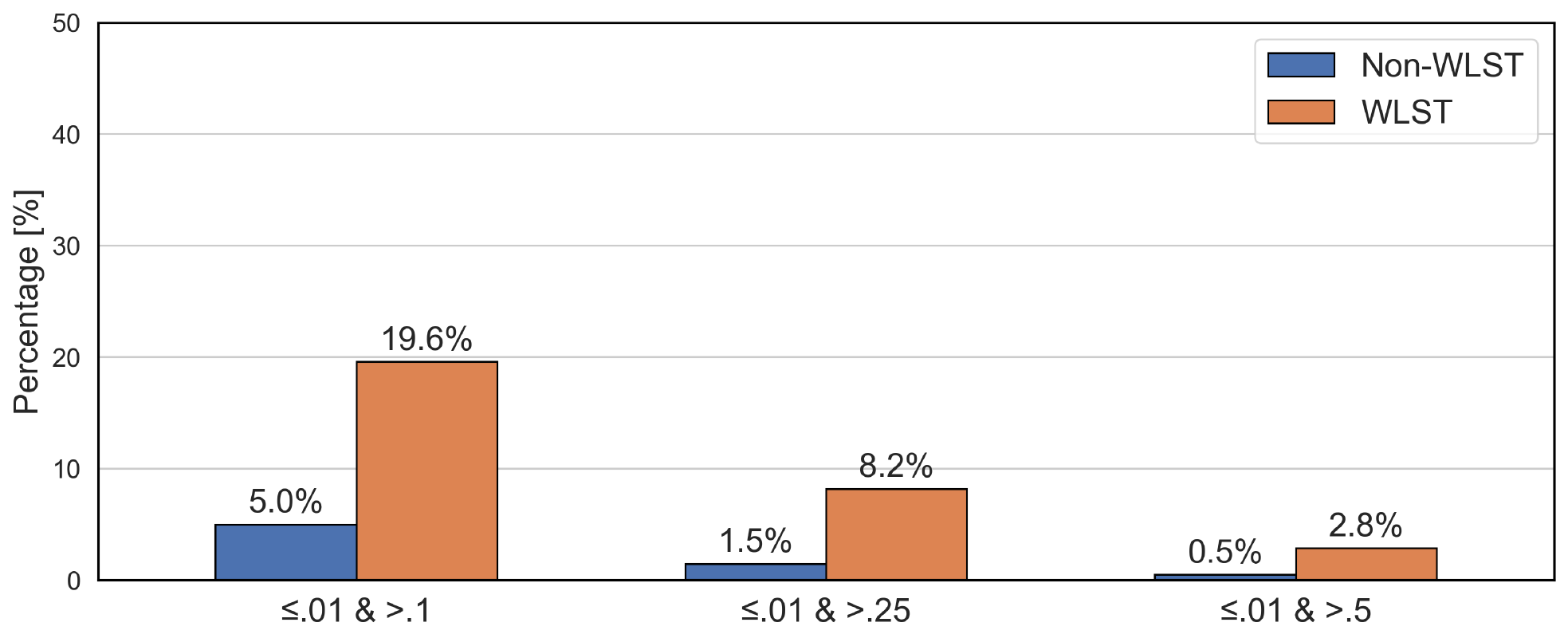}
    \caption{Percentage of non-WLST (blue) and WLST (orange) patients for whom at least one model predicts a likelihood of recovery of $\leq$1\% and at least one other of $>$10\% (left), $>$25\% (middle), or $>$50\% (right).}
    \Description{Percentage of non-WLST and WLST patients with significant discrepancies between different models' predictions. Discrepancies are higher for WLST patients.}
    \label{fig:barplots-comparison}
\end{figure*}

\Cref{fig:boxplots-comparison} shows the distribution of the differences between the most optimistic and the most pessimistic recovery prognosis of any of the ten models $M_{obs},...,M_{exp_{agr_w}+ip}$ for a given patient (left) as well as the variance of different models' predictions (right) for patients with known labels (\textit{Non-WLST}) and those with unknown labels (\textit{WLST}).
We see that both the maximum differences and variance in predictions are larger for WLST patients.
\Cref{fig:barplots-comparison} shows the percentage of patients for whom at least one model predicts a likelihood of recovery of $\leq$1\% and at least one other model predicts $>$10\% (left), $>$25\% (middle), or $>$50\% (right), indicating stark discrepancies.
Again, these discrepancies are significantly more pronounced for WLST patients.
Taken together, this means that the heterogeneity resulting from different modeling approaches is particularly strong for WLST patients.
This is especially problematic because those are precisely the instances for which we do \emph{not} have true labels and, as a result, cannot assess the correctness of predictions.

\end{document}